# Costate-focused models for reinforcement learning

Bita Behrouzi[1], Xuefei Liu[1], and Douglas Tweed[1,2]

[1] Department of Physiology, University of Toronto, Toronto, ON M5S 1A8, Canada
[2] Centre for Vision Research, York University, Toronto, ON M3J 1P3, Canada

Many recent algorithms for reinforcement learning are model-free and founded on the Bellman equation. Here we present a method founded on the costate equation and models of the state dynamics. We use the costate — the gradient of cost with respect to state — to improve the policy and also to "focus" the model, training it to detect and mimic those features of the environment that are most relevant to its task. We show that this method can handle difficult time-optimal control problems, driving deterministic or stochastic mechanical systems quickly to a target. On these tasks it works well compared to deep deterministic policy gradient, a recent Bellman method. And because it creates a model, the costate method can also learn from mental practice.

Research in reinforcement learning has shown the effectiveness of model-free algorithms founded on the Bellman equation and action-value functions (also known as $Q$-functions) or closely related quantities such as value- or advantage functions [1-6]. Here we present a different approach based on models and the costate equation of optimal control [7, 8], which may work better in some tasks.

As usual in reinforcement learning, the setting involves an *agent* in an *environment* which evolves through time according to a rule or function $f$, called the *state dynamics*. For instance, the agent might be a brain and the environment its body, in which case $f$ might represent the mechanics of that body. If $s_t$ is the *state* of the environment at time $t$, and $a_t$ is the *action* taken by the agent at this time (say, the motor commands issued by the brain), then at the next time step, $t + \Delta t$, the state takes a new value

$$(1) \quad s_{t+\Delta t} = s_t + \Delta s_t = s_t + \Delta t f(s_t, a_t)$$

The agent chooses its actions based on a function $\mu$ called its *policy*

$$(2) \quad a_t = \mu(s_t)$$

At each time step, the agent receives feedback about the quality of its performance: a reward signal $r$ or cost-rate $c$, which may depend on $s$ and $a$

$$(3) \quad c(s_t, a_t)$$

For instance, in the vestibulo-ocular reflex, which counterrotates the eyes when the head moves, so as to keep the visual image stable, the cost-rate is a neural signal coding retinal-image slip [9]. In reaching, it might be some function of the distance from hand to target. The aim is to learn a policy that minimizes cost-rates through time.

## Learning with costates

Here we explore an approach we call *costate policy gradient*, derived from algorithms of Parisini and Zoppoli [10] and Saerens et al. [11]. It is an interesting option for several reasons.

Rather than learning $Q$, it breaks the task into separate and possibly easier pieces, learning the state dynamics, $f$, and (in some versions of the algorithm) the cost-rate function, $c$. These functions, $f$ and $c$, can be acquired by supervised learning, which may be faster and more reliable than the bootstrapping [1] used in many $Q$-based methods. Further, $Q$-functions can be complex, calling for large networks. And while $f$ may also be complex, the costate equation suggests a simple way that the model can be focused on the most relevant aspects of the environment.

Also, costate methods allow an agent to improve by mental practice, using its internal model of the environment. The brain seems have such models, as humans can predict the sensory consequences of actions, and form plans based on imagined scenarios.

### Costate policy gradient

We consider episodic, or in other words finite-time or finite-horizon tasks, where the aim is to minimize the *cost C* of a movement, which is the time-integral of the cost-rates throughout the motion, from time 0 to a final time $T$,

$$(4) \quad C = \Delta t \sum_{t=0...T} c_t = \Delta t \sum_{t=0...T} c(s_t, \mu(s_t))$$

The policy $\mu$ is a multilayer network with adjustable parameters $\theta^\mu$, and so our aim is to adjust $\theta^\mu$ to reduce the average cost $E[C]$ over some repertoire of motions, e.g. reaches from a variety of initial states. To make those adjustments, we perform a lot of motions, and after each one we compute the gradient of its cost $C$ with respect to each of its actions $a_t$, from $t = 0$ to $T$. To find that gradient, we note that $a_t$ can affect $C$ in 2 ways, by altering $c_t$ and $s_{t+\Delta t}$, and therefore by the chain rule,

(5) $\qquad \partial C/\partial a_t = \Delta t [\partial c_t/\partial a_t + \partial C/\partial s_{t+\Delta t} \, \partial f/\partial a_t]$

This formula shows that we need $\partial C/\partial s_{t+\Delta t}$ to get $\partial C/\partial a_t$. To find the $\partial C/\partial s_t$, we again apply the chain rule,

(6)
$$\begin{aligned}\partial C/\partial s_t &= \Delta t [\partial c_t/\partial s_t + \partial c_t/\partial a_t \, \partial \mu/\partial s_t] + \\ &\quad \partial C/\partial s_{t+\Delta t} \, ds_{t+\Delta t}/ds_t \\ &= \Delta t [\partial c_t/\partial s_t + \partial c_t/\partial a_t \, \partial \mu/\partial s_t] + \\ &\quad \partial C/\partial s_{t+\Delta t} [I + \Delta t (\partial f/\partial s_t + \partial f/\partial a_t \, \partial \mu/\partial s_t)] \\ &= \partial C/\partial s_{t+\Delta t} + \Delta t [\partial c_t/\partial s_t + \partial c_t/\partial a_t \, \partial \mu/\partial s_t + \\ &\quad \partial C/\partial s_{t+\Delta t} (\partial f/\partial s_t + \partial f/\partial a_t \, \partial \mu/\partial s_t)]\end{aligned}$$

Hence if we know (or can estimate) the functions $f$, $c$, and $\mu$, we can compute the final $\partial C/\partial s$,

(7) $\qquad \partial C/\partial s_T = \Delta t [\partial c_T/\partial s_T + \partial c_T/\partial a_T \, \partial \mu/\partial s_T]$

and then sweep back in time, using (6) to compute all the $\partial C/\partial s_t$ in turn down to $\partial C/\partial s_{\Delta t}$. In control theory the $\partial C/\partial s_t$ are called *costates*, and (6) is the *costate equation* [8].

We plug these $\partial C/\partial s_t$ into (5) to find all the $\partial C/\partial a_t$, and use those derivatives to improve the policy, adjusting $\theta^\mu$ down the gradient

(8) $\qquad \partial C/\partial \theta^\mu = \Sigma_{t=0...T} \partial C/\partial a_t \, \partial a_t/\partial \theta^\mu$

That is, we backpropagate the $\partial C/\partial a_t$ through the $\mu$ network [12].

### Learning $f$ and $c'$

Costate policy learning begins with a brief stage of *motor babbling* [13], where the agent tries random combinations of states and actions, and observes the resulting $\Delta s$'s, to learn the state dynamics $f$ by backprop based on the error signal

(9) $\qquad e^f = \Delta t \langle f \rangle (s_t, a_t) - \Delta s_t$

Also during the babble stage, the agent can learn to estimate the cost-rate $c$, or some convenient related function. That is, normally in reinforcement learning we provide the signals $c_t$ to an agent that knows its job is to optimize their sum or time-integral [1]; but it can be more efficient to provide a related variable $c'_t$ to an agent that knows its job is to minimize the integral of $\varphi(c'_t)$ for some specified function $\varphi$. We give an example below, under *Tests*, subsection *Cost-rate and c'*.

### Learning the policy

After the babble stage, the agent starts doing *rollouts* — full movements from random initial states. For each time point $t$ in each rollout, the agent computes $\partial C/\partial a_t$ using (5). There are then several ways it might use those derivatives to improve its policy, and we will look at 2 of them. In the *direct* method, we store all the $\partial C/\partial a_t$ and use them to adjust the parameters $\theta^\mu$ of $\mu$, using (8), after the rollout is complete. In the *indirect* method, we apply the $\partial C/\partial a_t$ immediately, at each time step in the rollout, to adjust the parameters $\theta^{\mu^-}$ of a network $\mu^-$ called the *shadow* policy (because it is a copy of $\mu$ but does not affect the state). Before the rollout stage, we set $\mu^- = \mu$, and then after each rollout, we nudge $\mu$'s weights and biases 10% of the way toward those of $\mu^-$, and set $\mu^- = \mu$ again.

### Focusing the model

In many tasks, the environment is complex but not all aspects of it are equally relevant. That fact shows up clearly in equations (5) and (6), where the state-dynamics function $f$ never appears on its own, but is always multiplied by the costate, $\partial C/\partial s_{t+\Delta t}$, meaning all that matters about $f$ is its projection onto that vector. Therefore in the rollout stage we can focus the model $\langle f \rangle$ by adjusting it, based no longer on the error (9) but on

(10) $\qquad e = \langle \partial C/\partial s_{t+\Delta t} \rangle [\Delta t \langle f \rangle (s_t, a_t) - \Delta s_t]$

When we supplement the costate policy gradient (CPG) method with the focusing mechanism (10), we call the result *costate-focus* (CF) learning.

We find that CF usually works better with indirect policy updates (via the shadow policy $\mu^-$), and CPG with direct. And for CF, it is useful to gate policy-adjustment based on the accuracy of the dynamics model $\langle f \rangle$, as measured by the model error $e$ (10). That is, we adjust $\theta^{\mu^-}$ only in time steps where the normalized squared error (the mean of $e^2$ divided by the within-minibatch variance of $\langle \partial C/\partial s_{t+\Delta t} \rangle \Delta s_t$) is < 1.

### Pseudocode for costate-focus learning

*// Babble stage*
```
for minibatch = 1 to n_b
  s = 2 (rand(n_s, n_m) − 0.5)
  a = 2 (rand(n_a, n_m) − 0.5)
  adjust ⟨ c' ⟩ using e^c' = ⟨ c' ⟩(s, a) − c'(s, a)
  adjust ⟨ f ⟩ using e^f = Δt (⟨ f ⟩(s, a) − f(s, a))
end
```

```
μ⁻ = μ  // create shadow policy

// Rollout stage
for rollout = 1 to $n_{rolls}$

  // Forward sweep
  $s_0$ = 2 (rand($n_s$, $n_m$) – 0.5)  // initial states
  for t = 0 to T in steps of Δt
    $a_t$ = μ($s_t$)
    $s_{t+\Delta t}$ = $s_t$ + Δt f($s_t$, $a_t$)
  end

  // Backsweep
  ⟨$c_T$⟩ = φ(⟨ c′ ⟩($s_T$, $a_T$))
  backprop 1 – ⟨$c_T$⟩² through ⟨ c′ ⟩ to get ⟨∇c⟩
  ⟨∂C/∂$a_T$⟩ = Δt ⟨∂$c_T$/∂$a_T$⟩
  compute ⟨∂C/∂$s_T$⟩ using (7)
  for t = T – Δt to 0 in steps of –Δt
    e = ⟨∂C/∂$s_{t+\Delta t}$⟩[Δt ⟨f⟩($s_t$, $a_t$) – Δ$s_t$]
    adjust ⟨f⟩ using e
    ⟨$c_t$⟩ = φ(⟨ c′ ⟩($s_t$, $a_t$))
    backprop 1 – ⟨$c_t$⟩² through ⟨ c′ ⟩ to get ⟨∇c⟩
    backprop ⟨∂C/∂$s_{t+\Delta t}$⟩ through ⟨f⟩ to get ⟨∂C/∂$s_{t+\Delta t}$ ∇f⟩
    compute ⟨∂C/∂$a_t$⟩ using (5)
    if e is small …
      backprop ⟨∂C/∂$a_t$⟩ through μ⁻ to adjust it
    compute ⟨∂C/∂$s_t$⟩ using (6)
  end
  μ ← μ + τ(μ⁻ – μ)
  μ⁻ ← μ

end
```

In this paper, $n_m$ = 100, Δt = 0.1, T = 3, φ = tanh.

**Tests**

We compared CPG and CF with a recent Bellman algorithm, *deep deterministic policy gradient* (DDPG) [4], on *time-optimal* tasks [14]: learning to move various mechanical systems quickly and accurately to a target. This is a challenging set of tasks, and an important one for the brain, which often has to move the eyes, head, or limbs rapidly from one posture to another.

In each task, the environment was a mechanical system defined by Δ$s_t$ = Δt f($s_t$, $a_t$). The function f was always second-order, like the laws of mechanics. That is, the state vector $s_t$, of dimensionality $n_s$, consisted of 2 subvectors, the *configuration* $q_t$ and the *velocity* $v_t$, each of dimensionality $n_q$ = $n_s$/2, and the state dynamics took the form

(11)    Δ$s_t$ = Δ[$q_t$; $v_t$] = Δt [$v_t$; α($s_t$, $a_t$)]

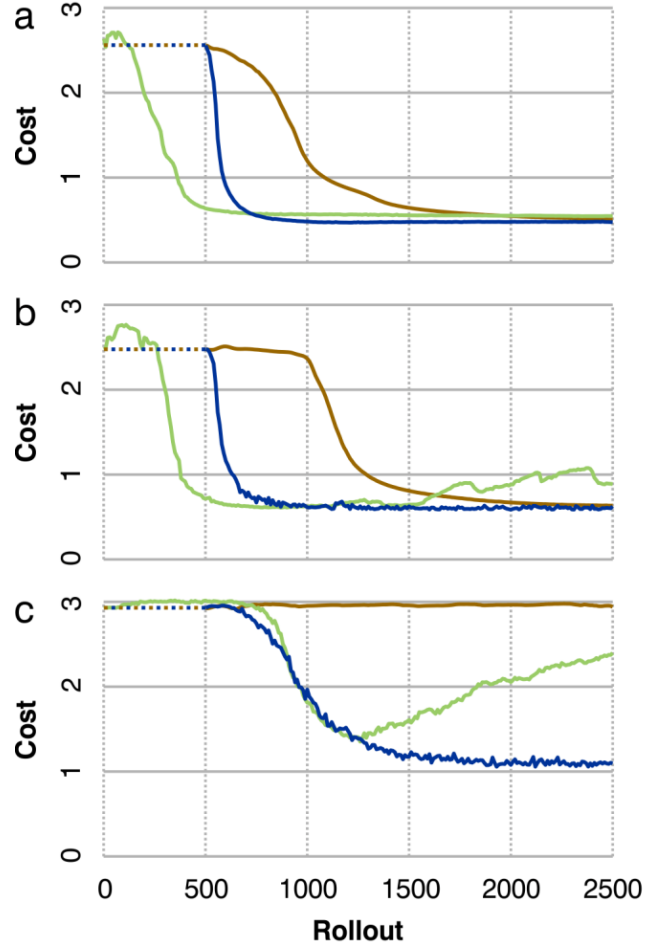

**Figure 1.** Learning curves for CF (blue), CPG (gold), and DDPG (green) on time-optimal tasks with linear state dynamics, each curve an average over 10 tasks. Horizontal dotted lines indicate that CF and CPG began with babble stages equivalent to 500 rollouts (see text under *Learning curves*). **a**) Simple tasks where $n_s$ = 10, $n_c$ = 1, $n_C$ = 4, $n_μ$ = 314, and $n_{est}$ = 4483 (for CF and CPG) or 4501 (for DDPG). **b**) Harder tasks where $n_s$ = 30, $n_c$ = 1, $n_C$ = 4, $n_μ$ = 554, and $n_{est}$ = 1507 (for CF and CPG) or 1528 (for DDPG). **c**) Still harder tasks, where $n_s$ = 100, $n_c$ = 2, $n_C$ = 8, $n_μ$ = 3124, and $n_{est}$ = 3661.

where the function α was the acceleration. In other words, $a_t$ affected the change in only the second part of the state vector; the change in the first part, $q_t$, was determined by $v_t$. To get a varied set of acceleration functions, we computed α($s_t$, $a_t$) using randomly selected linear functions or 3-layer tanh nets.

The neural network ⟨f⟩, which *learned f*, was a 3- or 4-layer *relu* net, *not* constrained to be second-order. That is, the agent did not know, at the outset, that the state dynamics had the form (11).

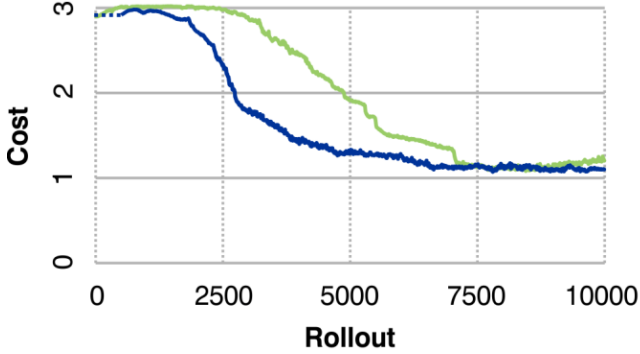

**Figure 2.** Tasks with very small estimator and policy networks, where $n_s = 100$, $n_c = 2$, $n_C = 8$, $n_\mu = 444$, and $n_{est} = 954$ (for CF) or 941 (for DDPG). Colors as in Figure 1. CF used a babble stage of up to 14 000 minibatches.

### Cost-rate and $c'$

The cost-rate was

(12) $$c_t = \tanh(s_t^\top B s_t)$$

where $B$ was a diagonal matrix of non-negative elements. This $c$ function encouraged the agent to move quickly to the target state, $s = 0$ [14].

We defined

(13) $$c' = s_t^\top B s_t, \quad \varphi = \tanh$$

A network $\langle c' \rangle$ learned to estimate $c'$ by backprop, during the babble stage, using the error signal

(14) $$e^{c'} = \langle c' \rangle(s_t, a_t) - c'_t$$

We set $B$ in (12) so that $c_t$ depended on only some elements of the state $s_t$, just as, in real life, only some aspects of your surroundings matter to you. We defined $n_c$ to be the number of elements of $s_t$ that affected $c_t$, and we set the first $n_c$ elements on the diagonal of $B$ to 10, and all other elements to 0.

Even if the current cost-rate, $c_t$, depended on only the first $n_c$ elements of $s_t$, many more elements of $s_t$ might affect future values of $s_1, s_2, \ldots, s_{nc}$, and therefore future $c$'s and the total cost, $C$. We defined $n_C$ to be the number of elements of $s_t$ that affected the cost, and we structured the $\alpha$ function of (11) to ensure that no other state element influenced $C$. The dimensionality of the action vector, $n_a$, was always $n_C/2$.

| Method | $n_s$ | $n_c$ | $n_C$ | $n_\mu$ | $n_{est}$ | $C_{min}$ | $C_{final}$ |
|---|---|---|---|---|---|---|---|
| DDPG | 10 | 1 | 4 | 314 | 4501 | 0.47 | 0.47 |
| CPG | 10 | 1 | 4 | 314 | 4483 | 0.51 | 0.51 |
| CF | 10 | 1 | 4 | 314 | 4483 | 0.46 | 0.48 |
| DDPG | 30 | 1 | 4 | 554 | 1528 | 0.59 | 0.89 |
| CPG | 30 | 1 | 4 | 554 | 1507 | 0.63 | 0.61 |
| CF | 30 | 1 | 4 | 554 | 1507 | 0.58 | 0.60 |
| DDPG | 100 | 2 | 8 | 3124 | 3661 | 1.14 | 2.38 |
| CPG | 100 | 2 | 8 | 3124 | 3661 | 2.87 | 2.96 |
| CF | 100 | 2 | 8 | 3124 | 3661 | 1.01 | 1.09 |
| VCF | 100 | 2 | 8 | 3124 | 3652 | 0.93 | 0.92 |
| DDPG | 100 | 2 | 8 | 444 | 941 | 1.01 | 1.21 |
| CF | 100 | 2 | 8 | 444 | 954 | 0.99 | 1.10 |
| VCF | 100 | 2 | 8 | 444 | 940 | 0.90 | 0.93 |

**Table 1.** Comparison of CF, CPG, DDPG, and VCF (vector costate-focus, described below) on tasks in linear, deterministic environments.

### Networks

All learning networks had relu neurons in their hidden layers, and all had linear ones in their output layers, except $\mu$, which used tanh to bound its outputs. The networks $\langle c' \rangle$, $\langle Q \rangle$, and $\mu$ always had 4 layers. We defined $n_{est}$ to be the number of adjustable parameters in all the estimator networks available to an agent, not counting DDPG's $Q'$, i.e. for DDPG, $n_{est}$ was the number of parameters in $\langle Q \rangle$, while for CF and CPG it was the number in $\langle f \rangle$ and $\langle c' \rangle$ together. $n_\mu$ was the number of adjustable parameters in the policy.

### Blocks

We ran tests in *blocks* of 10 trials each. Each *trial* presented a new task, with a new environment, and ran for 2000–10 000 rollouts, each rollout being a minibatch of $n_m = 100$ movements. In each trial, all the methods involved in the test — DDPG, CPG, or CF — learned the same task, with the same environment, initial policy, and set of 100 test movements. Initially and after every 10 rollouts we ran each method's current policy on the 100 test movements, and recorded its cost, averaged across those movements.

### Hyperparameters

We updated network weights and biases using Adam [15], with the standard values for its β hyperparameters, 0.9 and 0.999. Its third parameter, the learning-rate constant η, was set as described below.

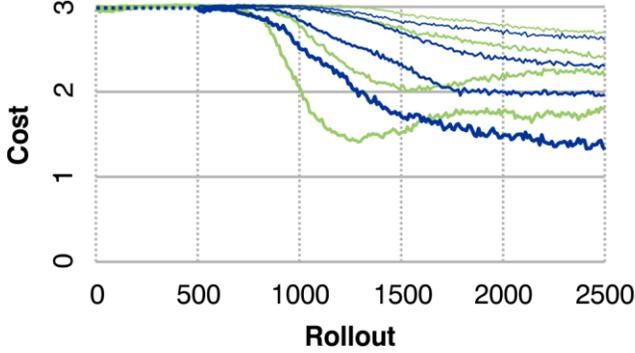

**Figure 3.** Learning in stochastic environments. Here $n_s = 100$, $n_c = 2$, $n_C = 8$, $n_\mu = 3124$, and $n_{est} = 3661$, as in Figure 1c, but now Gaussian noise has been added to the acceleration, with standard deviation 10 for the thickest curves, through 20 and 30 to 40 in the thinnest curves.

For CPG, the hyperparameters were $\eta_b$ (used to adjust $\langle f \rangle$ and $\langle c' \rangle$ in the babble stage), $\eta_\mu$ (to adjust the policy), and $\eta_{c'}$ (to improve $\langle c' \rangle$ during the rollout stage with a replay buffer of off-policy data, in the same way that DDPG adjusts $\langle Q \rangle$). Their values were $\eta_b = 0.001$, $\eta_{c'} = 0.0003$, and $\eta_\mu = 0.0003$.

For CF, the hyperparameters were $\eta_b$ (used to adjust $\langle f \rangle$ and $\langle c' \rangle$ in the babble stage), $\eta_f$ (to focus $\langle f \rangle$ during the rollout stage), $\eta_\mu$ (to adjust the policy), and $\tau$ (to nudge $\boldsymbol{\mu}$ toward the shadow policy $\boldsymbol{\mu}^-$). Their values were $\eta_b = 0.001$, $\eta_f = 0.0001$, $\eta_\mu = 0.001$, and $\tau = 0.1$, except in the tasks in Figures 2 and 6, where $\eta_f = 0.0003$.

DDPG had 3 hyperparameters that we adjusted to optimize performance: the learning-rate constants $\eta_Q$ and $\eta_\mu$ (used to adjust $\langle Q \rangle$ and $\boldsymbol{\mu}$) and $\tau$ (to nudge DDPG's target networks $Q'$ and $\boldsymbol{\mu}'$ toward $\langle Q \rangle$ and $\boldsymbol{\mu}$). We set $\eta_Q = 0.0003$, $\eta_\mu = 0.0001$, and $\tau = 0.0003$ except for the tasks in Figure 2, where $\eta_Q = 0.001$, $\eta_\mu = 0.0001$, and $\tau = 0.00003$. All of DDPG's other hyperparameters were set as in the paper that introduced the method [4].

### Learning curves

To show learning curves for each method, we plotted mean costs on the test set versus rollout. For analysis (but not for the plots), we smoothed each curve with a running average, i.e. we replaced each data point in the curve with the mean of the 5 most recent points. We found the lowest-cost point in each smoothed curve, took the mean of those lowest costs across the 10 curves in a block, and called that average value $C_{min}$ — the minimum cost achieved by that method on that block of tasks. Similarly, we recorded the *final* cost in each smoothed curve, took the mean, and called it $C_{final}$.

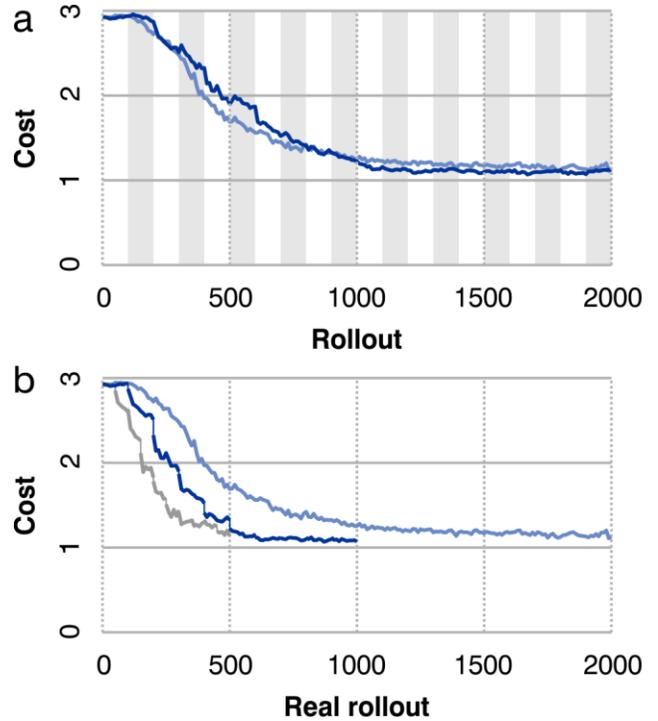

**Figure 4.** CF can learn by mental practice. Tasks as in Figure 1c, though here each trial began with a babble stage of just 3000 minibatches. **a)** Even when half the rollouts (grey bands) were imaginary, the learning curve (dark blue) was about as good as when all the rollouts were real (pale blue). **b)** Here the 2 curves from the top panel are replotted vs their real (non-imaginary) rollouts. The grey curve is a third experiment where ¾ of the rollouts were imaginary.

For CF and CPG, learning began with a babble stage using minibatches of 100 examples each. Therefore thirty such minibatches contained as many examples of states, actions, and cost-rates as a single 30-time-step rollout. So if there were, say, 15 000 babble minibatches, we regarded them as equivalent to 15 000 / 30 = 500 rollouts, and when we plotted CF's and CPG's learning curves together with DDPG's, we shifted the CF and CPG curves 500 rollouts to the right, omitted their final 500 rollouts, and indicated the babble stage with a horizontal dotted line, and as in Figure 1. For DDPG, $C_{min}$ was then the minimum cost achieved within its 2500 rollouts, whereas for CF and CPG it was the minimum achieved within 2500 – 500 = 2000 rollouts. But for all 3 methods — CF, CPG, and DDPG — we defined $C_{final}$ to be the cost after all 2500 rollouts, to give all the methods equal time to become unstable. So for CF and CPG, it sometimes happened that $C_{final} < C_{min}$.

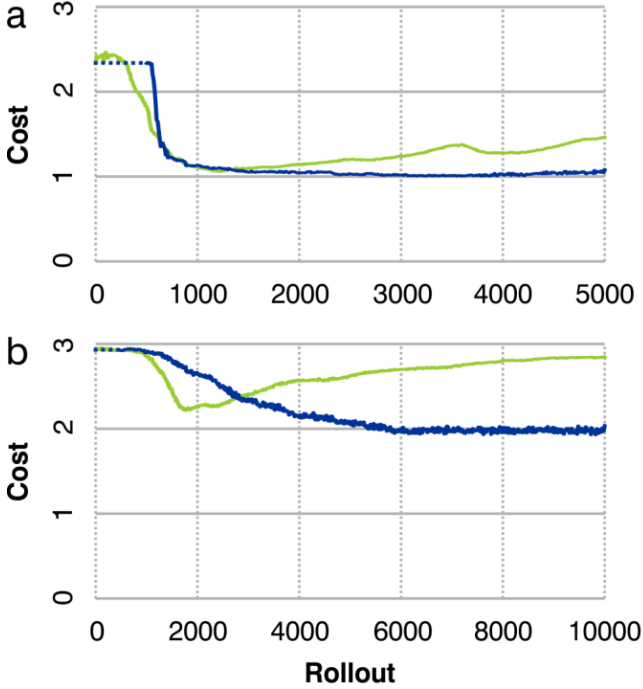

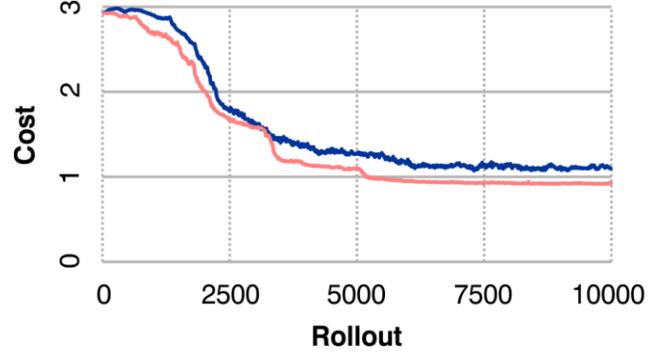

**Figure 5.** Learning curves for CF (blue), and DDPG (green) on tasks with nonlinear state dynamics. **a)** Tasks where $n_s$ = 30, $n_c$ = 1, $n_C$ = 4, $n_\mu$ = 554, and $n_{est}$ = 2822 (for CF) or 2815 (for DDPG). **b)** Harder tasks, where $n_s$ = 100, $n_c$ = 2, $n_C$ = 8, $n_\mu$ = 3124, and $n_{est}$ = 3656 (for CF) or 3661 (for DDPG).

**Figure 6.** VCF (pink) compared with CF (blue) on the tasks from Figure 2. For VCF, $n_{est}$ = 940.

### Results

All 3 methods worked well when the state dynamics were simple and the agents had adequate capacity in their estimator networks (i.e. large enough $n_{est}$) to approximate $f$ and $c'$, or $Q$, as in Figure 1a. In harder tasks, DDPG showed instability and CPG eventually failed, as shown in Figures 1b and 1c.

Both CF and DDPG were able to learn even with very small estimator networks, though only given very long learning times, as shown in Figure 2. Test results from Figures 1 and 2 (and 5 and 6) are summarized in Table 1.

CF and DDPG coped about equally well with stochastic dynamics, but CF was more stable, as shown in Figure 3.

Unlike DDPG and other Bellman methods, CF creates an internal model $\langle f \rangle$ of its environment, and so can learn by mental practice using that model (Figure 4a). As a result, it can improve with less real experience (Figure 4b).

Figures 1–4 show tasks with linear state dynamics, but the results were much the same when the dynamics were nonlinear. In Figure 5 the acceleration functions (defined by 3-layer tanh networks) were non-affine in both $s$ and $a$, though still differentiable.

### Vector reinforcement

If the agent knows not just the cost-rate $c$ but also its gradient with respect to $s$ and $a$, then it can use that exact gradient vector rather than an estimate of it derived from a $\langle c' \rangle$ network. This method, called *vector costate-focus* or VCF, usually outperformed CF, as in Figures 6 and 7.

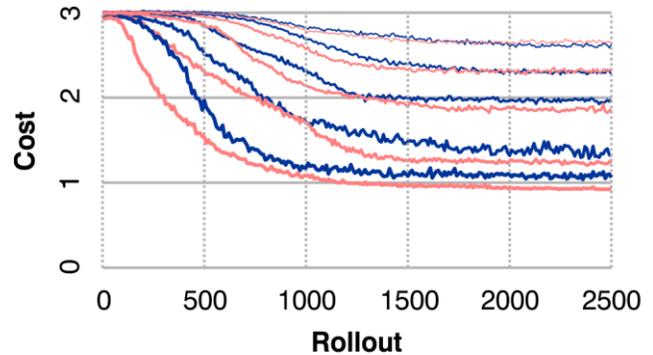

**Figure 7.** VCF (pink) compared with CF (blue) on the tasks from Figures 1c and 3.

So both forms of costate-focus learn well in complex environments, by creating models that mirror the task-relevant aspects of the state dynamics.

Acknowledgments. We thank Ali Mrani Alaoui, Sara Scharf, and Collin Wilson for comments. This work was supported by NSERC grant 391349-2010.